\documentclass[letterpaper, 10 pt, conference]{ieeeconf}  

\IEEEoverridecommandlockouts                              

\usepackage{graphics} 
\usepackage{epsfig} 
\usepackage{times} 
\usepackage{amsmath} 
\usepackage{amssymb}  
\usepackage{bbding}
\usepackage{booktabs}
\usepackage{algorithm}
\usepackage{algpseudocode}
\usepackage{pifont}

\usepackage{tabularx}
\usepackage{subcaption}
\DeclareMathOperator*{\argmax}{argmax}

\title{\LARGE \bf
Multi-Modal Legged Locomotion Framework with Automated Residual Reinforcement Learning
}

\author{Chen Yu and Andre Rosendo
\thanks{Chen Yu and Andre Rosendo are with School of Information Science and Technology, ShanghaiTech University, Shanghai, China
        {\tt\small \{yuchen, arosendo\}@shanghaitech.edu.cn}}%
}

\begin{document}

\maketitle
\thispagestyle{empty}
\pagestyle{empty}

\begin{abstract}
While quadruped robots usually have good stability and load capacity, bipedal robots offer a higher level of flexibility / adaptability to different tasks and environments. A multi-modal legged robot can take the best of both worlds. 
In this paper, we propose a multi-modal locomotion framework that is composed of a hand-crafted transition motion and a learning-based bipedal controller---learnt by a novel algorithm called Automated Residual Reinforcement Learning. 
This framework aims to endow arbitrary quadruped robots with the ability to walk bipedally.
In particular, we 1) design an additional supporting structure for a quadruped robot and a sequential multi-modal transition strategy; 2) propose a novel class of Reinforcement Learning algorithms for bipedal control and evaluate their performances in both simulation and the real world.
Experimental results show that our proposed algorithms have the best performance in simulation and maintain a good performance in a real-world robot.
Overall, our multi-modal robot could successfully switch between biped and quadruped, and walk in both modes. Experiment videos and code are available at \url{https://chenaah.github.io/multimodal/}.

\end{abstract}

\section{Introduction}

The development of quadruped robots is an interesting and popular research field in robotics since they are simpler than hexapod robots while having better stability and load capacity than biped robots \cite{he2019survey}. Many promising quadruped robots have been developed in the last few years, such as BigDog \cite{bigdog}, Spot \cite{spot}, Mini Cheetah \cite{Bosworth2015}, and ANYmal \cite{anymal}. These robots are implemented with bionic structures to benefit from the quadruped performance from animals. 

On the other hand, bipedal robots are also of interest to researchers due to their humanoid structures and hence the ability to mimic agile human locomotion \cite{Saeedvand2019}. In comparison to quadruped robots, humanoid robots usually have a higher level of flexibility and adaptability to different environments and applications. Current examples of well known bipedal robots are Atlas from Boston Dynamics \cite{Kuindersma2016} and Cassie \cite{gong2019feedback}.

Motivated by the advantages of both quadruped and bipedal robots, we propose a novel multi-modal locomotion strategy that can enable a quadruped robot to walk in bipedal mode.
This locomotion strategy is composed of a novel hand-crafted transition mechanism and a learning-based bipedal controller---trained by a new algorithm called Automated Residual Reinforcement Learning (ARRL).
We implement this framework on a quadruped robot Mini Cheetah \cite{Bosworth2015}. The robot therefore could walk quadrupedally following the same MPC strategy proposed by Carlo et al. \cite{mitmpc} and walk bipedally with our ARRL algorithm
, remaining the same body structure and actuators for both locomotion modes. 

\textbf{Multi-Modal Locomotion.} Fukuda et al. \cite{fukuda2012multi} design a multi-locomotion robot, Gorilla Robot, inspired by male gorillas that can perform bipedal walking, quadruped walking, and brachiation. They propose a method called Passive Dynamic Autonomous Control and a corresponding gait selection strategy \cite{Kobayashi2013LocomotionSS, kobayashi2015selection}. Huang et al. \cite{huang2018design} design a crawling strategy for a biped walking with a rigid-flexible waist using CPG control. Earlier works involve controlling humanoid robots to crawl with hands and feet \cite{nishiwaki2007experimental, kamioka2015dynamic} or hands and knees \cite{sanada2009passing, kanehiro2007whole, kanehiro2004locomotion, huang2019dynamic}. However, these robots are either mechanically complicated or energy-consuming. For instance, the 22kg Gorilla Robot is powered by 24 AC motors of 20-30W. In addition, 
some of them walk relatively slowly and unstably because of their application of moving in narrow spaces.

Rather than designing a multi-locomotion robot from scratch, we propose a lightweight and low-cost solution to transit an off-the-shelf quadruped robot from quadruped mode to bipedal mode: we use a mechanical structure to provide a supporting polygon for bipedal mode, which does not affect the quadruped locomotion controller. A static action sequence is designed to enable the robot to switch smoothly between two modes. 

\textbf{Bipedal Locomotion Control.} The locomotion control for multi-modal robots is more challenging than single-modal robots because of the potential conflict of the mechanical and dynamics model for each mode of locomotion. The difficulty of converting a quadruped robot into a biped comes from 1) a higher limitation on load capacity per leg---as the full weight falls in two limbs---and 2) the small foot area with a limited degree of freedom of the foot to accommodate the potentially conflicting configuration space of the quadruped locomotion. 

Control of a bipedal robot is usually designed analytically based on a simplified dynamics model, such as Linear Inverted Pendulum. Once the robot is kinematics consistent, physically feasible trajectories can be calculated within a zero-moment-point (ZMP) constraint \cite{vukobratovic1969contribution}. Adjusting the ZMP dynamically during locomotion can make the walking process more robust \cite{kim2006experimental, kaneko2002design}. However, these hand-crafted modelling usually results in brittle and inaccurate controllers in complicated control problems that involve friction and contacts \cite{johannink2019residual}.

Another popular class of control algorithms are learning-based, such as Reinforcement Learning (RL). RL algorithms can search for a policy without any assumption on the dynamics model. It is shown that model-free RL is able to solve complicated 3D humanoid control tasks \cite{haarnoja2018soft, schulman2017proximal, Ziyu2017}. Johannink et al. \cite{johannink2019residual} propose Residual RL that combines hand-engineered and RL controllers: the final control policy is a superposition of a conventional feedback controller and a residual policy solved with RL. This improves the sample efficiency since the agent does not have to learn the structure of the control problem from scratch.

In our work, we wish to have an efficient learning-based controller that is also based on a few dynamics assumptions ---so that this whole multi-model framework is more scalable and could be easily applied to other quadruped robots.
Therefore, we propose a variant of Residual RL, Automated Residual RL (ARRL): the conventional feedback controller is still a component of the final control policy, but is also automatically trained by an optimiser: the conventional feedback controller and the RL agent are trained simultaneously. In this way, we do not have to hand-tune even the basic controller, while remaining the structure of the control problem. We show that ARRL can outperform both pure RL and pure parameter black-box optimisers. 





The contribution of this paper is 1) the design of a multi-modal legged locomotion strategy, 2) the proposal of a control algorithm ARRL, and 3) the analysis of the ARRL framework with respect to different kinematic primitives, black-box optimisers, and RL agents.






\section{Multi-Modal Transition Strategy}
\label{sec:trans}
Our multi-modal locomotion framework mainly contains two parts: a hand-crafted multi-modal transition strategy and a learning-based bipedal controller. The novel transition strategy makes multi-modal locomotion possible and leads to a unique control problem.

\subsection{Mechanical Design}
\begin{figure}[t]
     \centering
     \framebox{\parbox{3in}{
      \centering
     \begin{subfigure}[b]{0.21\textwidth}
         \centering
         \includegraphics[width=\textwidth]{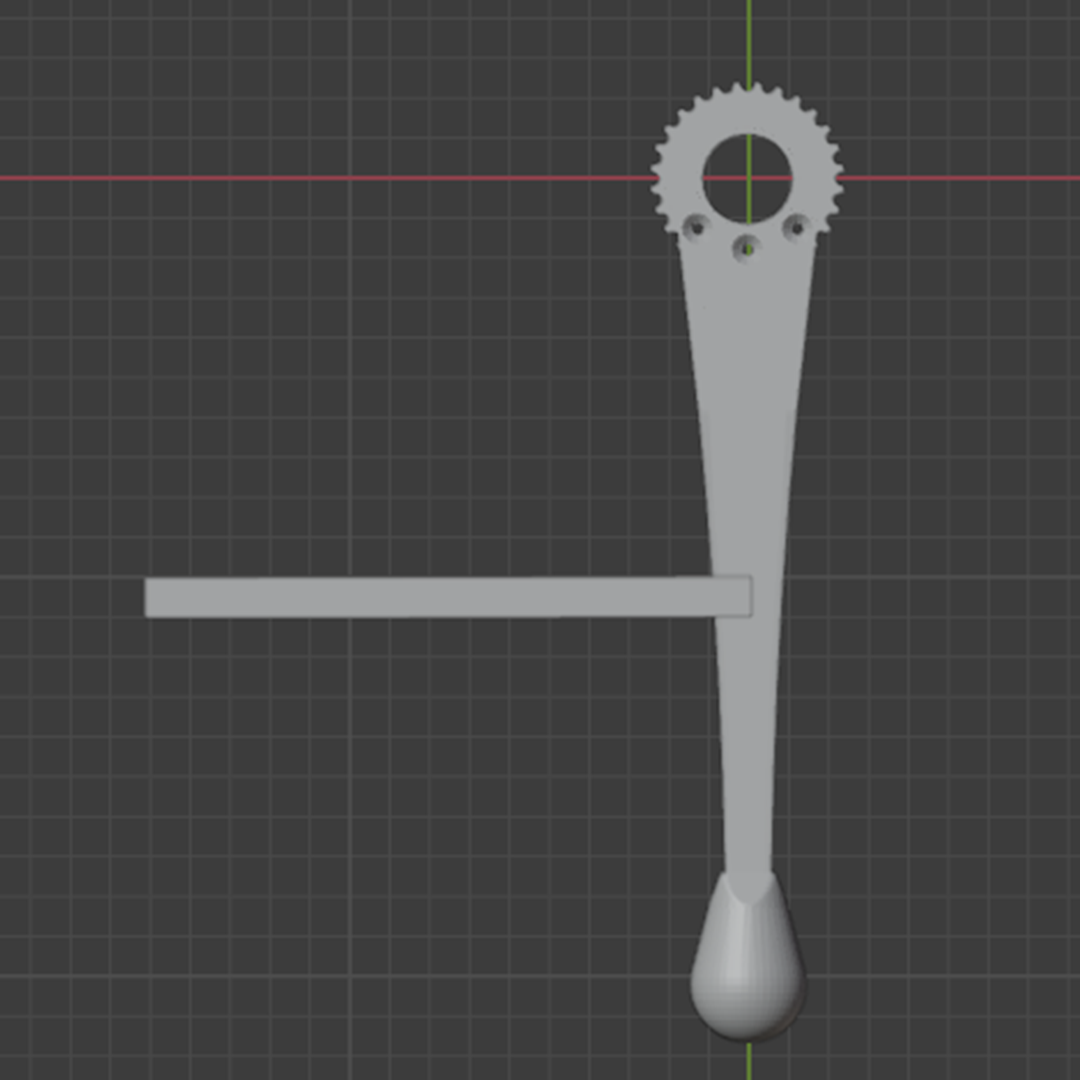}
     \end{subfigure}
     \hfill
     \begin{subfigure}[b]{0.21\textwidth}
         \centering
         \includegraphics[width=\textwidth]{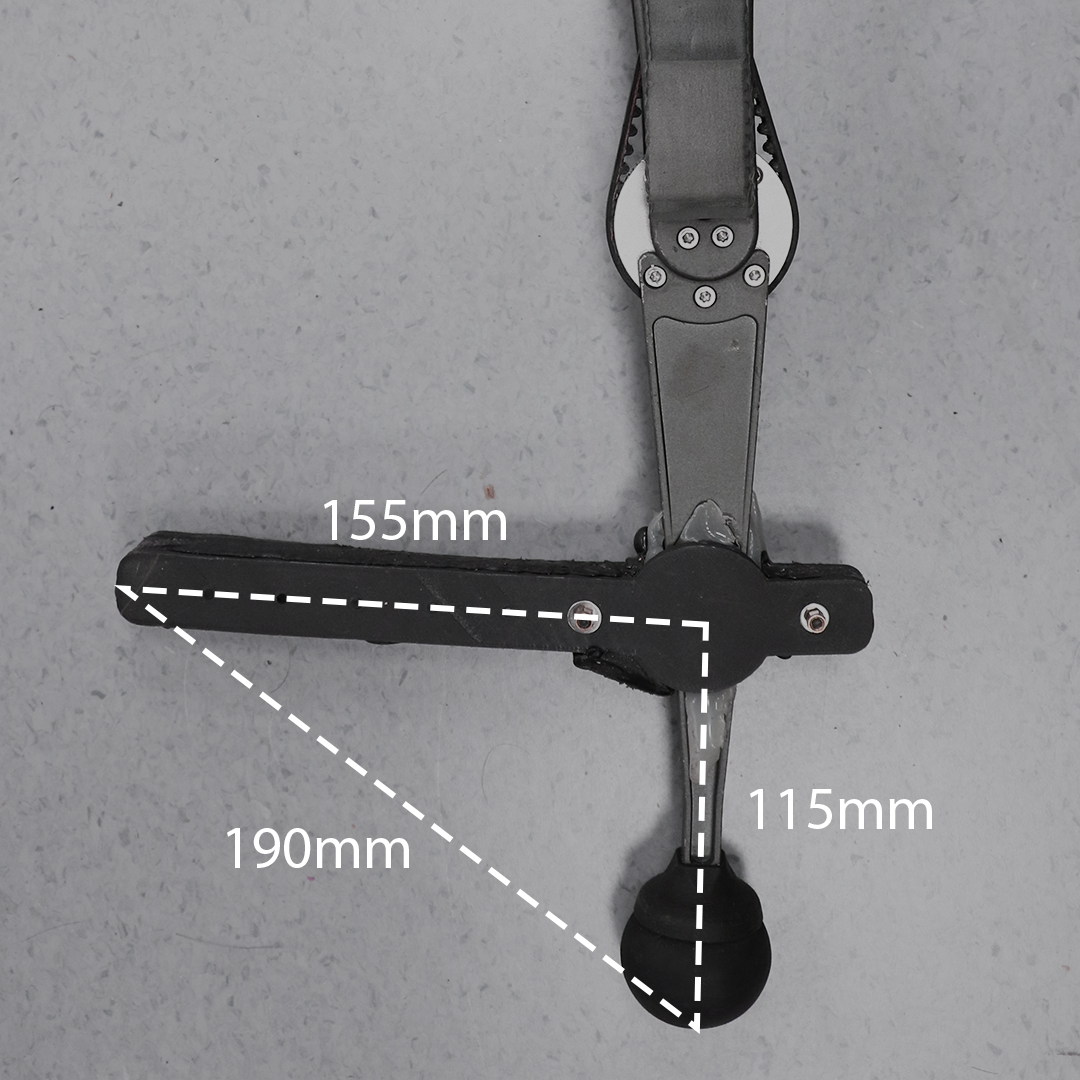}
     \end{subfigure}
     }}
        \caption{\textbf{A demonstration of the installed supporting stick.}
        We show the mechanical structure in the 3D design software (left) and the real robot (right).}
        \label{fig:feet}
\end{figure}

We design a 3D-printed stick and installed it on the shank of the hind legs of the original quadruped robot. The dimensions are designed to make sure that the robot has a) enough configuration space for quadruped locomotion in its quadruped mode and b) enough supporting convex polygons area for robust locomotion control in its bipedal mode. The design is illustrated in Fig. \ref{fig:feet}.




\subsection{Multi-Modal Transformation}

\begin{figure}[t]
      \centering
      \framebox{\parbox{3in}{
      \centering
      \includegraphics[width=0.38\textwidth]{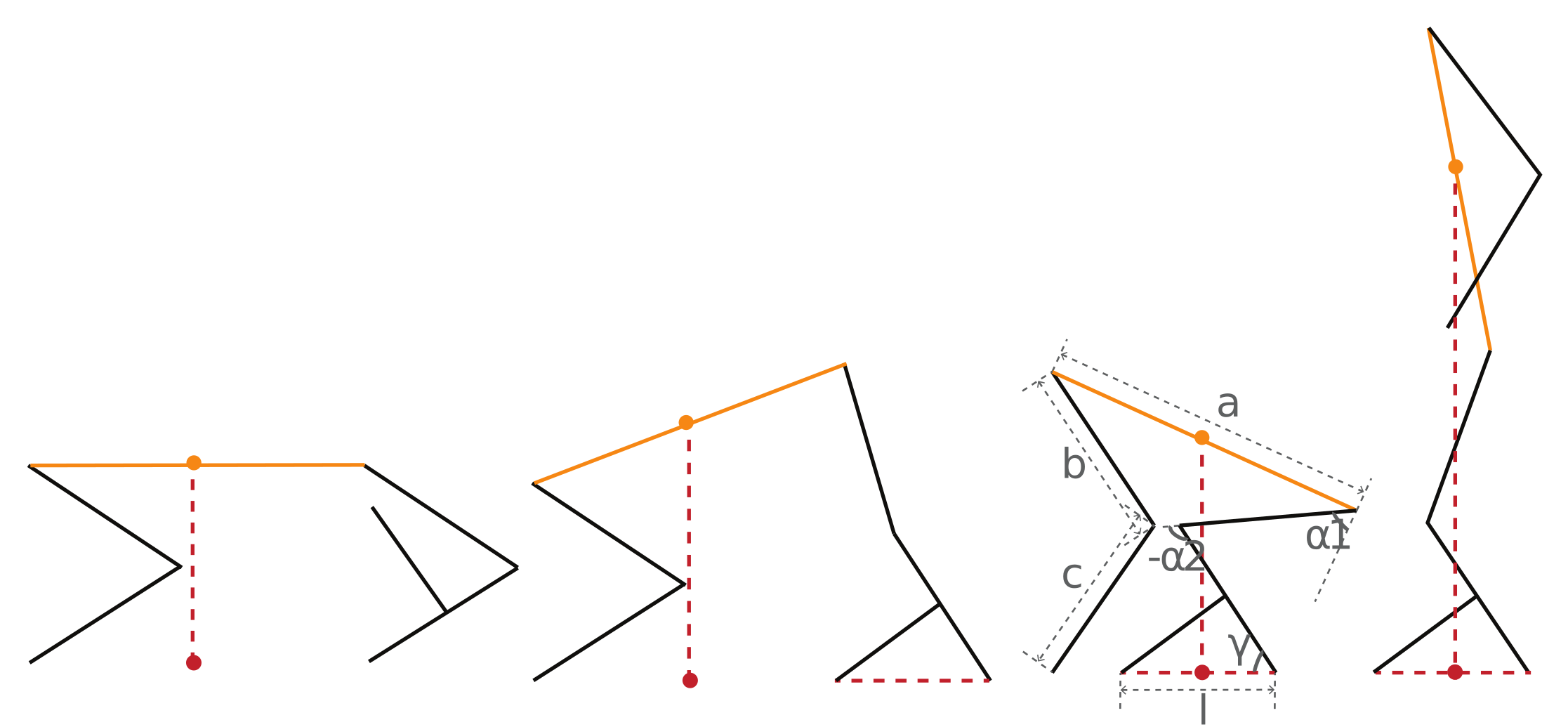}
      }}
      
      \caption{\textbf{An 2D illustration of the action sequence for the transition from quadruped mode to bipedal mode. }
      The orange line stands for the torso, the grey lines stand for legs, and the red lines demonstrate the projection of the CoM of the robot. During the transformation between the third state and the fourth, the CoM of the robot is always projected on the centre of the equivalent feet.}
      \label{fig:standing}
   \end{figure}
  
The action sequence of multi-modal transformation is demonstrated in Fig. \ref{fig:standing}. The mode transition starts from a state as a normal quadruped robot where the added supporting structure will not affect its quadruped locomotion. The hind legs of the robot will then bend to a position that all the legs and the supporting sticks are touching the floor. The hands of the front feet of the robot then move toward the hind legs horizontally, which is done by calculating the inverse kinematics (IK). Finally, in the standing phase of the transformation, the projection of the CoM of the robot on the floor is kept at the centre of the equivalent footprint. Denoting the position of the hip joint as $\alpha_1$ and position of knee joint as $\alpha_2$, we have an equation according to IK:
\begin{equation}
\alpha_1 = -\frac{\pi}{2} - \alpha_2 + \gamma -\arccos(\frac{(l/2 - b\cos(\alpha_2 - \gamma) - c \cos\gamma)}{ak}),
\label{eq:standing}
\end{equation}
where $a$ is the length of the torso, $b$ is the length of thigh, $c$ is the length of the shank, $\gamma$ is the angle between the shank and the ground when both the toe and the installed stick is touching the floor, and $k$ is a real number describing the position of the CoM of the robot projected on the torso. 

\section{Learning-Based Bipedal Control}
To solve the locomotion control problem of the robot in the bipedal mode, we propose and evaluate a variety of learning-based algorithms.

\subsection{Preliminaries}
The control problem can be modelled as a {continuous state space} Markov Decision Process (MDP), written as a tuple $M = (\mathcal{S}, \mathcal{A}, r, \mathcal{P}, \gamma, d_0)$, where $\mathcal{S}\in\mathcal{R}^{d_s}$ and $\mathcal{A}\in\mathcal{R}^{d_a}$ are the sets of states and actions; $r_t(\mathbf{s}_t, \mathbf{a}_t, \mathbf{s}_{t+1})$ is the reward function describes the objective of the task. $\mathcal{P}(\cdot|\mathbf{s}_t, \mathbf{a}_t)$ is the transition probability distribution, $\gamma \in [0, 1)$ is the discounting factor, and $d_0$ is the distribution of the initial state . The objective of an RL algorithm is to learn a policy that maximises the total rewards within an episode.

\subsection{Off-Policy Reinforcement Learning}
\label{sec:rl}
RL methods usually estimate the Q-value, which recursively evaluates the expected discounted return after taking action $\mathbf{a}$ at state $\mathbf{s}$. It can be described by the Bellman equation:
\begin{equation}
Q\left(\mathbf{s}_{t}, \mathbf{a}_{t}\right)=\mathcal{E}_{\mathbf{s}_{t+1}}\left[r_{t}+\gamma \max _{\mathbf{a}_{t+1}} Q\left(\mathbf{s}_{t+1}, \mathbf{a}_{t+1}\right)\right].
\label{eq:bellman}
\end{equation}
In this paper we specifically consider two state-of-the-art off-policy RL algorithms that optimise the Bellman equation by sampling off-policy transitions $(\mathbf{s}_t; \mathbf{a}_t; r_t; \mathbf{s}_{t+1})$ from a replay buffer.

\textbf{Soft Actor Critic (SAC).} SAC is a stochastic off-policy actor-critic reinforcement learning algorithm \cite{haarnoja2018soft}. It would maximise the entropy of policy besides the expected reward from the environment during training, which encourages state exploration of the agent.

\textbf{Twin Delayed Deep Deterministic Policy Gradients (TD3).} TD3 is a deterministic off-policy actor-critic reinforcement learning algorithm based on DDPG \cite{dankwa2019twin, Lillicrap2016ContinuousCW}. It uses two Q-functions, delayed policy update, and noise on the target action to improve the performance over baseline DDPG. 

In our problem, the state $\mathcal{S}$ include the roll, pitch, yaw, the angular velocities of the torso in the world frame along these three axes, and the eight motor angles (two for front abduction joints, four for all hip joints, and two for hind knee joints). For the action $\mathcal{A}$, we choose to use the position control mode of the actuators which is shown to be more controllable for RL \cite{peng2017learning}. It controls two front abduction joints, all four hip joints, and two hind knee joints. More specifically, the actions from the policy $\pi_{\theta}$ are served as an incremental angle of a joint after multiplying by a real number $k_a$. We design the reward function inspired by previous related work \cite{Schulman2016, tassa2012synthesis} as follows:
\begin{equation}
r = \mathrm{height} - 0.25\mathrm{max}(|\mathbf{a}_{\mathrm{front}}|) + \mathrm{distance} - \mathrm{cost}_l + 0.5,
\label{eq:reward}
\end{equation}
where $\mathbf{a}_{\mathrm{front}}$ represents the actions for the front legs and $cost_l = 0.1$ when an action is reaching the joint limit otherwise $cost_l = 0$. $0.5$ is a bonus for living. This reward function encourages the agent to walk as far as possible with the lowest possible energy consumption.
An episode ends when the $e_{pitch}$ in (\ref{eq:error}) $ > 0.65$ or the number of time steps reaches the maximum (1000). We score an episode with $T$ steps by $\sum_{t=0}^{T-1}r_t$.
In this paper, all the reward function, stop condition, and the scoring rule are consistent between simulation and real-world experiment.
\subsection{Residual Reinforcement Learning}
We use the strategy of residual RL \cite{johannink2019residual} to speed up the training process. The action $a_t$ is chosen by additively combining a model-free RL policy $\pi_{\theta}(\mathbf{a}_t | \mathbf{s}_t)$ and a simpler parametric policy $\pi_{\theta^\prime}(\mathbf{s}_t)$:
\begin{equation}
\mathbf{a}^{\prime}_t = \pi_{\theta}(\mathbf{s}_t) + \pi_{\theta^\prime}(\mathbf{s}_t).
\end{equation}

We assume a simple dynamics model and use two PD controllers, which serves as part of the basic controllers $\pi_{\theta^\prime}$ of our experiments, to utilise the front legs to compensate the movement of CoM. One PD controller to control the hip joints according to the pitch error $e_{pitch}(t)$ and the other to control the abduction joints according to the yaw error $e_{yaw}(t)$. The PD controller equation is 
\begin{equation}
u_{\text{hip}}=K_{p p} \cdot e_{\text{pitch}}(t) + K_{d p} \cdot \frac{d e_{\text{pitch}}(t)}{d t}
\label{eq:error}
\end{equation}
\begin{equation}
u_{\text{abduction}}=K_{p y} \cdot e_{\text{yaw}}(t) + K_{d y} \cdot \frac{d e_{\text{yaw}}(t)}{d t}.
\end{equation}
Here the referenced pitch is the pitch of the torso when the robot just finishes the multi-modal transformation described in (\ref{eq:standing}) and the referenced yaw is $0$. $u_{hip}$ and $u_{abduction}$ are the incremental angles of the front hip joints and abduction joints. $\{K_{p p}, K_{d p}, K_{p y}, K_{d y}\}$ is a subset of parameters $\pi^\prime$ for controller $\pi_{\theta^\prime}$.

Another part of the basic controller $\pi_{\theta^\prime}$ is an open-loop controller to generate gait patterns. Denote the position of hip and knee joints of a leg as $\mathbf{\alpha}=[\alpha_1, \alpha_2]$
We design four gaits as follows:

\textbf{Line.} We design a baseline gait that does not step forward, which changes the angles of the hip and knee joins in different direction with the same amplitude.

\textbf{Sine.} We design a gait trajectory that follows the pattern of half a period of sine function as follows:
\begin{equation}
\mathbf{\alpha} =
\begin{cases}
IK(x_0+\omega t\delta_x/\pi, y_0+A \sin(\omega t)), & t<\pi/\omega,\\
IK(x_0+\delta_x - \omega t\delta_x/\pi, y_0), & t \geq \pi/\omega
\end{cases}
\end{equation}

\textbf{Rose.} We also design a gait trajectory that follows the pattern of part of a rose function. We calculate $\mathbf{\alpha} = IK(x_t, y_t)$ by
\begin{equation}
x_t =
\begin{cases}
x_0+\delta_x \cos(2\alpha_p)\cos(\alpha_p), & t<\pi/\omega,\\
x_0+\delta_x - (t-\pi/\omega) \cdot b\delta_x/\pi, & t \geq \pi/\omega
\end{cases}
\end{equation}
\begin{equation}
y_t =
\begin{cases}
y_0+4A / \delta_x \cos(2\alpha_p)\sin(\alpha_p), & t<\pi/\omega,\\
y_0, & t \geq \pi/\omega
\end{cases}
\end{equation}
\begin{equation}
\alpha_p = 
\begin{cases}
\omega/4 \cdot (\pi/\omega-t), & t<\pi/\omega,\\
\omega/4 \cdot (2\pi/\omega-t), & t \geq \pi/\omega.
\end{cases}
\end{equation}

\textbf{Triangle.} This trajectory moves the toe upward, forward, and returning to the original spot, drawing a triangle pattern. We calculate $\mathbf{\alpha} = IK(x_t, y_t)$ by
\begin{equation}
x_t = 
\begin{cases}
x_0+\frac{\sin(2\omega t-\frac{\pi}{2})+1}{2} \cdot (x_1 - x_0), & t<\frac{\pi}{2\omega},\\
x_1-\frac{\sin(2\omega t-\frac{\pi}{2})+1}{2} \cdot (x_2 - x_1), &
\hfill \frac{\pi}{2b} \leq t < \frac{\pi}{\omega},    \\
x_0 + \delta_x - (t-\pi/\omega) \cdot \omega\delta_x/\pi, & t \geq \frac{\pi}{\omega}
\end{cases}
\end{equation}
\begin{equation}
y_t = 
\begin{cases}
y_0+\frac{\sin(2\omega t-\pi/2)+1}{2} \cdot (y_1 - y_0), & t<\pi/(2\omega),\\
y_0, & t \geq \pi/(2\omega).
\end{cases}
\end{equation}

\begin{figure}[t]
      \centering
      \framebox{\parbox{3in}{
      \centering
      \includegraphics[width=0.38\textwidth]{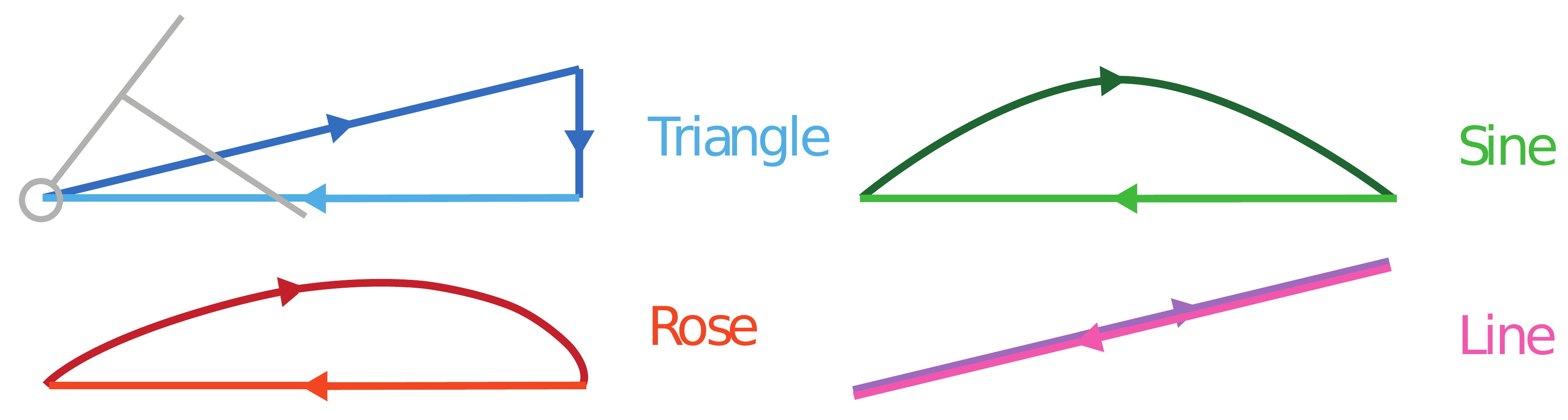}
      }}
      
      \caption{\textbf{Four gait patterns.} These are the kinematic primitives for an open-loop leg controller.}
      \label{fig:gait}
   \end{figure}

For all the gait patterns, two legs perform with a phase shift of half a period. An illustration of these gait patterns is shown in Fig. \ref{fig:gait}. 

The controllers for the front legs and hind legs described above constitute the basic controller $\pi_{\theta^{\prime}}$ for our experiment, where $\theta^{\prime} = \{A / \omega, \omega, K_{pp}, K_{dp}, K_{py}, K_{dy}, \delta_x\}$. The range for the parameter $A/\omega$ is $[8, 11]$ for the triangle gait and $[0, 3]$ for others. The ranges for $K_{pp}$, $K_{dp}$, $K_{py}$, and $K_{dy}$ are all $[0.00, 0.1]$. The range for $\delta_x$ is $[0, 0.05]$. All the parameters are randomly initialised before the training process.

\subsection{Evolution Strategy and Bayesian Optimisation}
\label{sec:bb}
Instead of hand-tuning the parameter $\theta^{\prime}$ for the additive parametric policy $\pi_{\theta^{\prime}}$, we consider a few black-box optimisers to find the optimum of these parameters during the training process.

\textbf{Covariance Matrix Adaptation Evolution Strategy (CMAES).} 
CMAES is an evolutionary algorithm with an ``evolutionary path'' storing the update direction of generations \cite{auger2012tutorial}. 
Previous works have shown its efficiency in control \cite{le2020sample, piergiovanni2019learning, chatzilygeroudis2017black}.

\textbf{Test-Based Population Size Adaptation (TBPSA).} TBPSA is a specific implementation of pcCMAES \cite{hellwig2016evolution}, a variant of CMAES. It evaluates points with a strong mutation rate and performs small steps in the best direction, relying on the longer-range trends of the objective landscape \cite{liu2020versatile}. 

\textbf{Bayesian Optimisation (BO).} BO is a global optimiser that uses observations to form the posterior distribution over the objective function \cite{snoek2012practical}.
It has been widely used in robot control due to its sample efficiency \cite{mailer2021evolving, guzman2020heteroscedastic, zhu2019bayesian}. 

These optimisers search for the optimal parameters $\theta^{\prime*}$ based on the training rewards (\ref{eq:reward}) averaged over a horizon of $H$ episodes:
\begin{equation}
    \theta^{\prime*} = \argmax_{\theta^{\prime}} \frac{1}{H} \sum_{n=0}^{H-1} \sum_{t=0}^{T-1}r_t.
\label{eq:argmax}
\end{equation}
Note that this evaluation is based on the returns from the training policy $\pi^{\prime}_{\theta}$ of an off-policy RL agent, which means that the optimisation process considers both exploration and exploitation of the RL agent. The proposed hybrid algorithm is shown in Alg. \ref{alg:main}.

\begin{algorithm}
\caption{Automated Residual Reinforcement Learning}
\begin{algorithmic}[1]
\Require RL policy $\pi_{\theta}$, basic parametric controller $\pi_{\theta^{\prime}}$
\While{$t_{\mathrm{total}} \leq t_{\mathrm{max}} $ total steps}
\For{$n = 0, \dots, H - 1 $ episodes}
\State Sample initial state $\mathbf{s}_0 \sim d_0$.
\For{$t = 0, \dots, T - 1 $ steps}
\State Obtain policy action $\mathbf{a}_t \sim \pi_{\theta}$.
\State Calculate action to execute $\mathbf{a}^{\prime}_t = \mathbf{a}_t + \pi_{\theta^{\prime}}(\mathbf{s}_t)$.
\State Transfer to next state $\mathbf{s}_{t+1} \sim \mathcal{P}(\cdot|\mathbf{s}_t, \mathbf{a}_t)$.
\State Store $(\mathbf{s}_t, \mathbf{a}_t, \mathbf{s}_{t+1})$ into the replay buffer $\mathcal{R}$.
\State Optimise $\theta$ with transitions $(\mathbf{s}, \mathbf{a}, \mathbf{s}) \sim \mathcal{R}$.
\EndFor
\State Calculate return $R_n=\sum_{t=0}^{T-1}r_t$ for episode $n$.
\EndFor
\State Optimise $\theta^{\prime}$ through (\ref{eq:argmax}).
\EndWhile
\end{algorithmic}
\label{alg:main}
\end{algorithm}

Overall, we use parameter optimisers (CMAES, TBPSA, or BO) to tune a conventional feedback controller \textit{simultaneously} with the training of the RL agent (SAC or TD3). This strategy can be viewed as using parameter optimisers to automatically tune the hyper-parameters of a residual RL method along with its own training process.
Since most RL methods (including SAC and TD3) are based upon the key assumption that the underlying Markov decision process is stationary, we assume that the transition dynamics are still fixed in a short period of time.


\subsection{Sim-to-Real Transfer}
\label{sec:simtoreal}
Beyond implementing the trained policy directly to the real robot, we gradually increased the parameter $A$ of the controller $\pi_{\theta^{\prime}}$ for each episode while satisfying the condition that errors $e_{\mathrm{pitch}}$ and $\frac{d e_{\mathrm{pitch}}}{d t}$, defined in (\ref{eq:error}), should always be below a threshold. We also progressively increased the weight of policy $\pi_theta$ until $1$. Specifically, we calculate the error based on the absolute values of $e_{\mathrm{pitch}}(t)$ and $\frac{d e_{\mathrm{pitch}}(t)}{d t}$, averaged over a period.
Denoting the two errors as $\mathbf{e}$, it yields:
\begin{equation}
A = 
\begin{cases}
k_1t, & k_1t < A_{\mathrm{set}} \quad \textrm{and} \quad  \mathbf{e} < \mathbf{e}_{\mathrm{threshold}} ,\\
A, & k_1t < A_{\mathrm{set}} \quad \textrm{and} \quad  \mathbf{e} \geq \mathbf{e}_{\mathrm{threshold}} ,\\
A_{\mathrm{set}}, & \mathrm{otherwise}
\end{cases}
\label{eq:prog0}
\end{equation}

\begin{equation}
a_t = 
\begin{cases}
\pi_{\theta}(s_t) + k_2t\pi_{\theta^\prime}(s_t), & k_2t < 1 \, \textrm{and} \,  \mathbf{e} < \mathbf{e}_{\mathrm{threshold}} ,\\
\pi_{\theta}(s_t) + k_2t^{\prime}\pi_{\theta^\prime}(s_t), & k_2t < 1 \, \textrm{and} \,  \mathbf{e} \geq \mathbf{e}_{\mathrm{threshold}} ,\\
\pi_{\theta}(s_t) + \pi_{\theta^\prime}(s_t), & \mathrm{otherwise},
\end{cases}
\label{eq:prog1}
\end{equation}
where $k_2t^{\prime}$ denotes the last updated $k_2t$. This method could help the closed-loop described in (\ref{eq:error}) to calculate a good upper arms positioning before the full actions of legs are performed, which greatly improves the balance and the sim-to-real transferability. 

\section{Results}

\subsection{Multi-Modal Transition}
\begin{figure*}[tb]
     \centering
     \framebox{\parbox{0.98\textwidth}{
       \centering
     \begin{subfigure}[b]{0.13\textwidth}
         \centering
         \includegraphics[width=\textwidth]{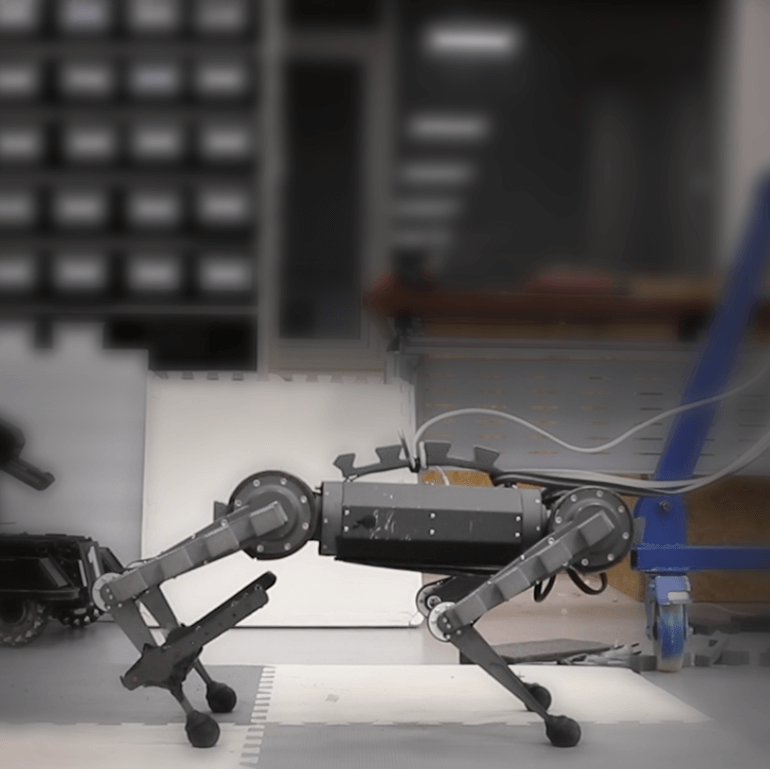}
     \end{subfigure}
     \hfill
     \begin{subfigure}[b]{0.13\textwidth}
         \centering
         \includegraphics[width=\textwidth]{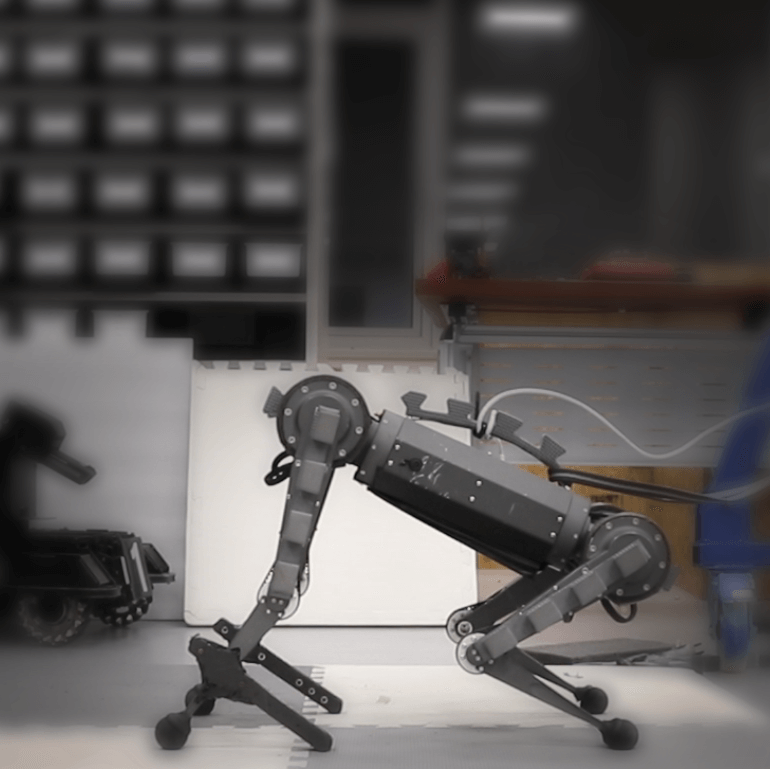}
     \end{subfigure}
     \hfill
     \begin{subfigure}[b]{0.13\textwidth}
         \centering
         \includegraphics[width=\textwidth]{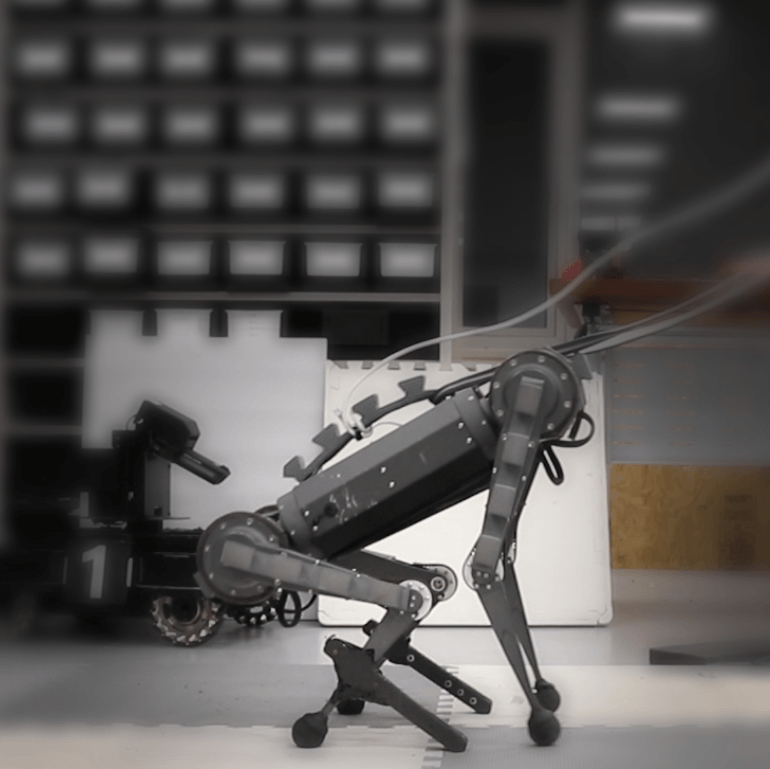}
     \end{subfigure}
     \hfill
     \begin{subfigure}[b]{0.13\textwidth}
         \centering
         \includegraphics[width=\textwidth]{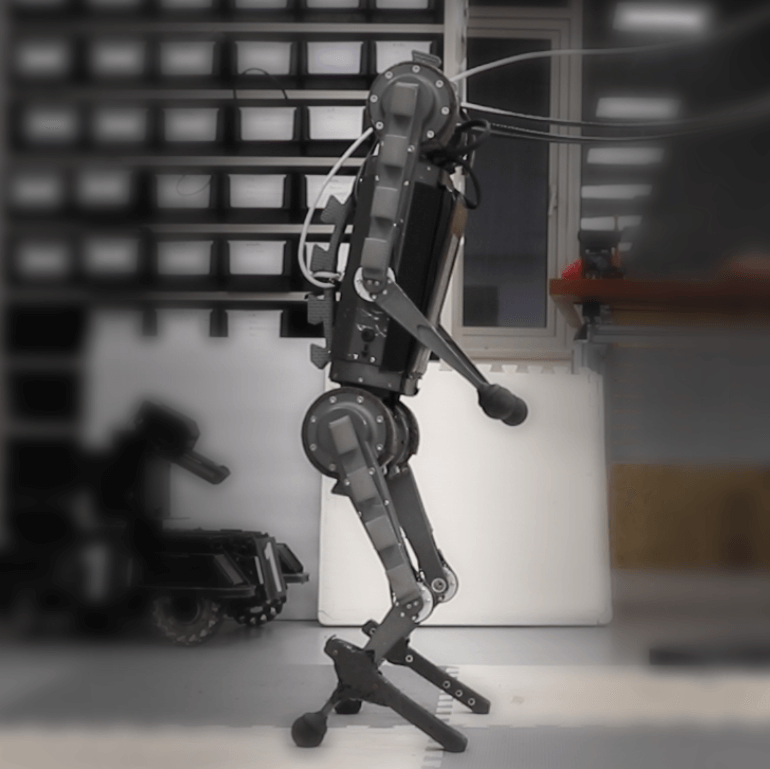}
     \end{subfigure}
     \hfill
     \begin{subfigure}[b]{0.13\textwidth}
         \centering
         \includegraphics[width=\textwidth]{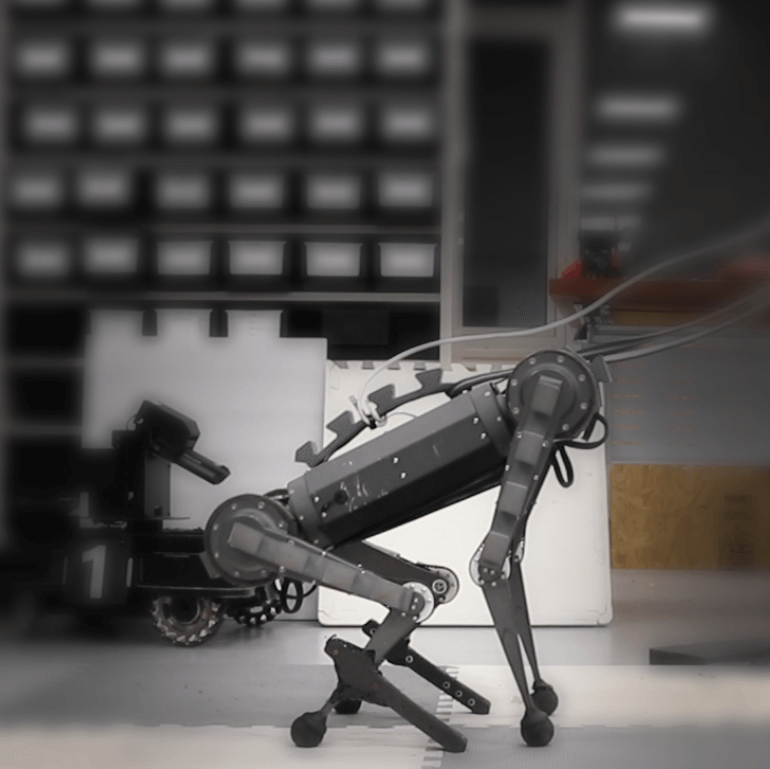}
     \end{subfigure}
     \hfill
     \begin{subfigure}[b]{0.13\textwidth}
         \centering
         \includegraphics[width=\textwidth]{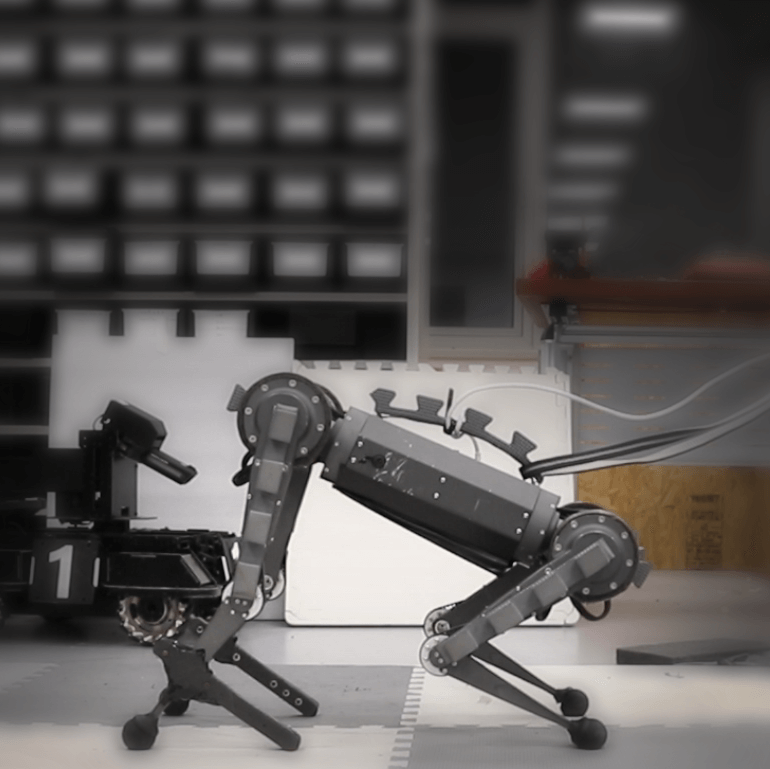}
     \end{subfigure}
     \hfill
     \begin{subfigure}[b]{0.13\textwidth}
         \centering
         \includegraphics[width=\textwidth]{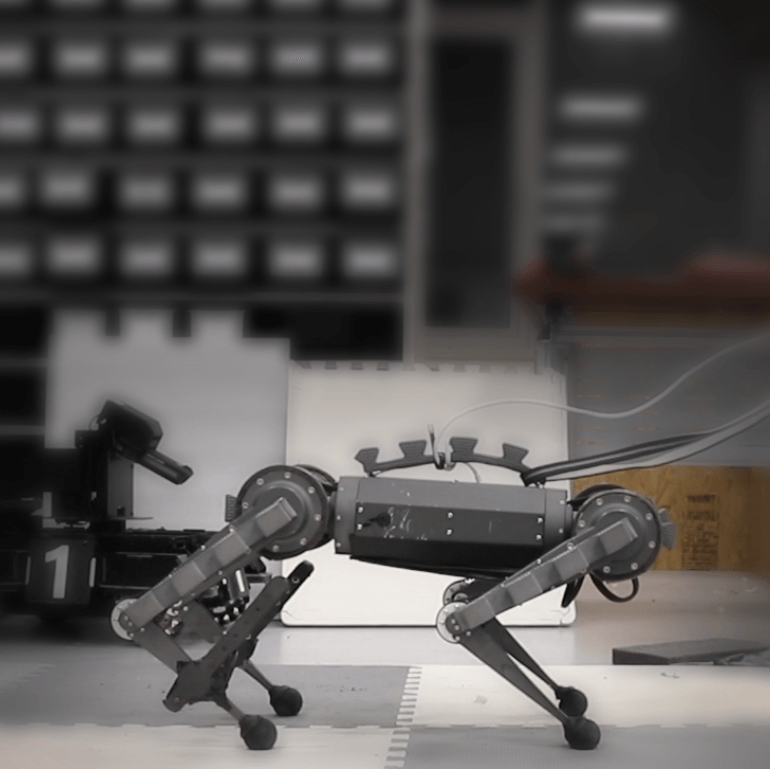}
     \end{subfigure}
     }}
        \caption{\textbf{Snapshots of the transition between bipedal and quadruped modes.} This is the first part of our proposed multi-modal locomotion framework. In the quadruped mode, the additional structure does not affect the quadruped locomotion controller, while in the bipedal mode the supporting structure successfully provides a stability polygon for the robot to keep bipedal balance. 
        }
        \label{fig:real_standing}
\end{figure*}

The multi-modal transition strategy described in Sec. \ref{sec:trans} and a reverse action sequence can both perform well on the real robot, as shown in Fig. \ref{fig:real_standing} and the supplementary video.

In the quadruped mode, the additional supporting structure does not affect the quadruped locomotion controller; in the bipedal mode, the supporting structure can successfully provide a supporting polygon for the robot to keep balance.
In this way, the robot can arbitrarily switch between these two locomotion modes. 

      

\subsection{Baseline for Bipedal Locomotion Control}
For this unique control task, we test two types of classic learning-based methods---using RL to learn a walking policy, and using black-box optimisers to tune a parametric controller---before the evaluation of our ARRL algorithm. In the following experiments, returns are calculated by accumulated rewards (\ref{eq:reward}) and all the results are averaged over three trials. For real-world implementation, we only consider the best policy learnt by the agent during the whole training process. 
Thus, for clarity, we show the best return obtained by the agent by each time step.

\begin{figure}[tb]
    \centering
    \framebox{\parbox{0.47\textwidth}{
    \centering
    \includegraphics[width=0.47\textwidth]{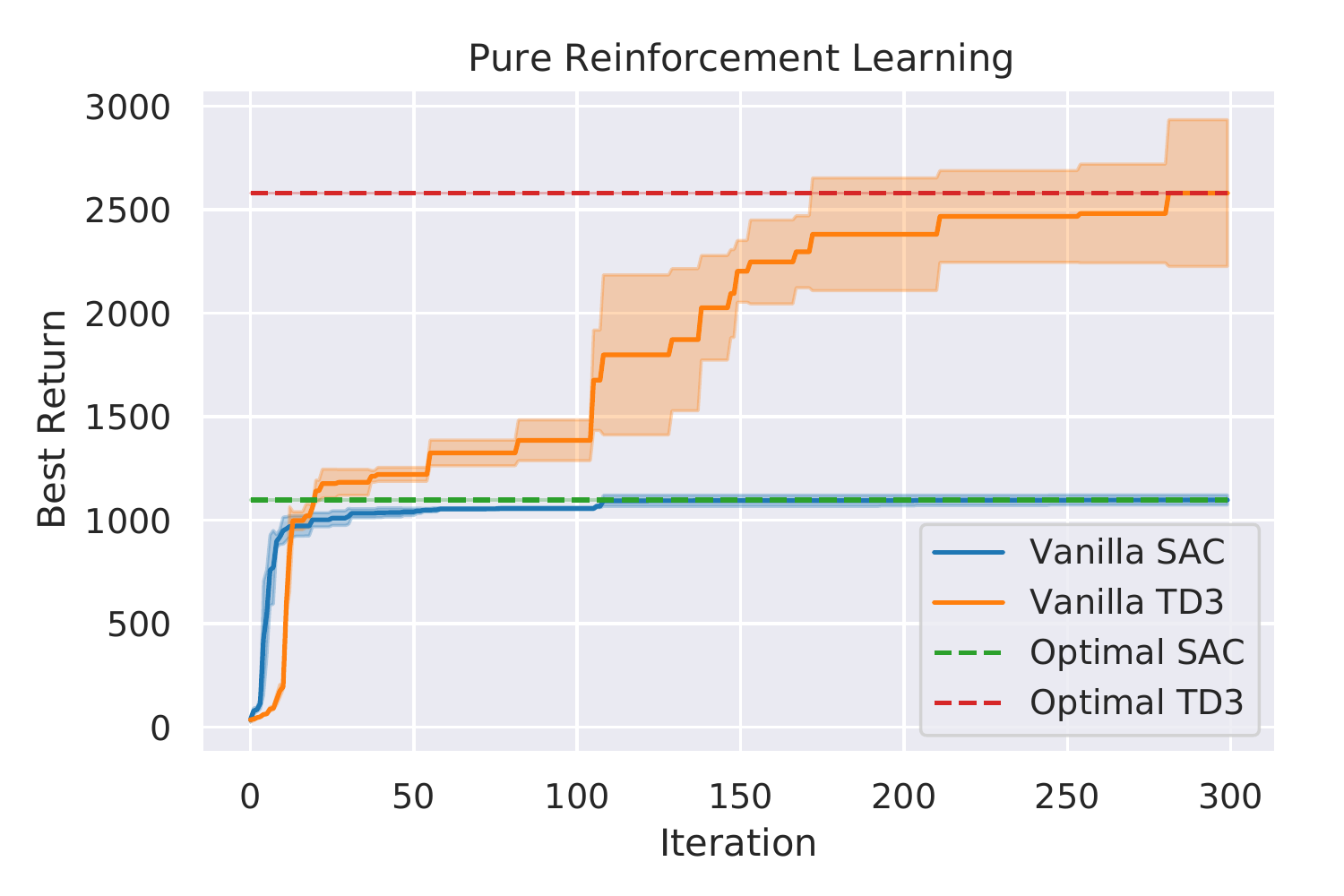}
    }}
    \caption{\textbf{Learning of bipedal locomotion using pure RL.} During the training process, the agent of SAC reaches an optimum score at $1097.23$ and the agent from TD3 obtains a maximum return of $2580.14$.}
    \label{fig:rl}
\end{figure}

Fig. \ref{fig:rl} shows the training curves of two standard RL agents, SAC and TD3. Here, we only use model-free RL algorithms introduced in Sec. \ref{sec:rl} to control the abduction joints of the front legs and all hips and knees joints. During the training process, SAC reaches an optimum score at $1097.23$ and the TD3 obtains a maximum return of $2580.14$. SAC seems to be stuck in a local maximum in this task.

\begin{figure*}[tb]
     \centering
     \framebox{\parbox{0.98\textwidth}{
       \centering
       \includegraphics[width=0.98\textwidth]{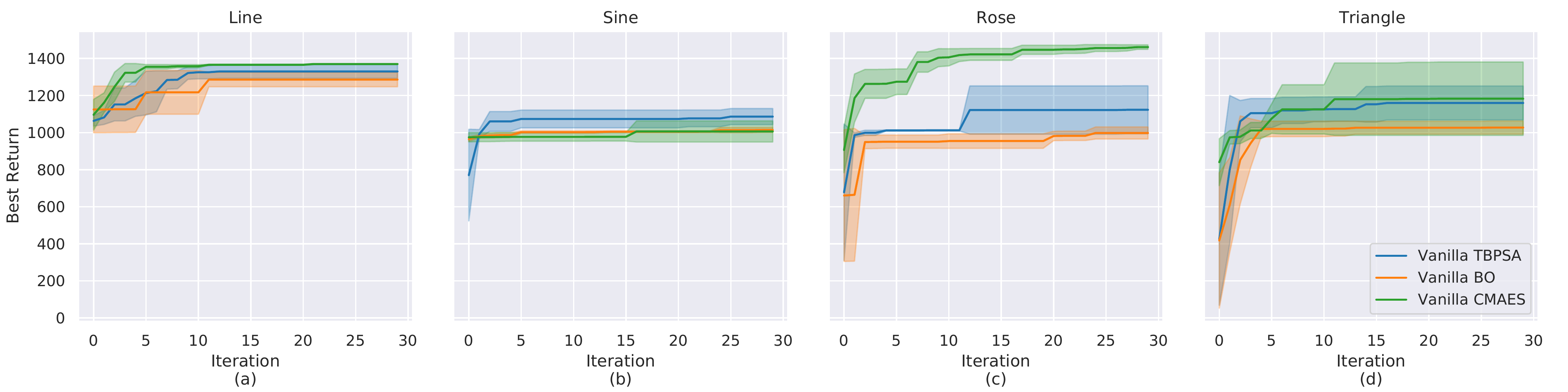}

     }}
        \vfill
        \caption{\textbf{Learning of bipedal locomotion using black-box optimisers.} All three methods can improve the controllers with four different motion primitives; and reach a return higher than SAC but lower than TD3. The Rose gait with CMAES reaches the highest score.}
        \label{fig:bb}
\end{figure*}

Fig. \ref{fig:bb} shows the training curves of three block-box optimisers: TBPSA, CMAES, and BO, with different motion primitives: Line, Sine, Rose, and Triangle. In this case, we only use parametric controller $\pi_{\theta^{\prime}}$ trained by block-box optimisers introduced in Sec. \ref{sec:bb}. All three methods can improve the controllers with four different motion primitives; and reach return higher than SAC but lower than TD3. With respect to algorithms, TBPSA and CMAES consistently outperform BO; with respect to motion primitives, the Rose gait reaches the highest score.

\subsection{ARRL for Bipedal Locomotion Control}

\begin{figure*}[tb]
     \centering
     \framebox{\parbox{0.98\textwidth}{
       \centering
       \includegraphics[width=0.98\textwidth]{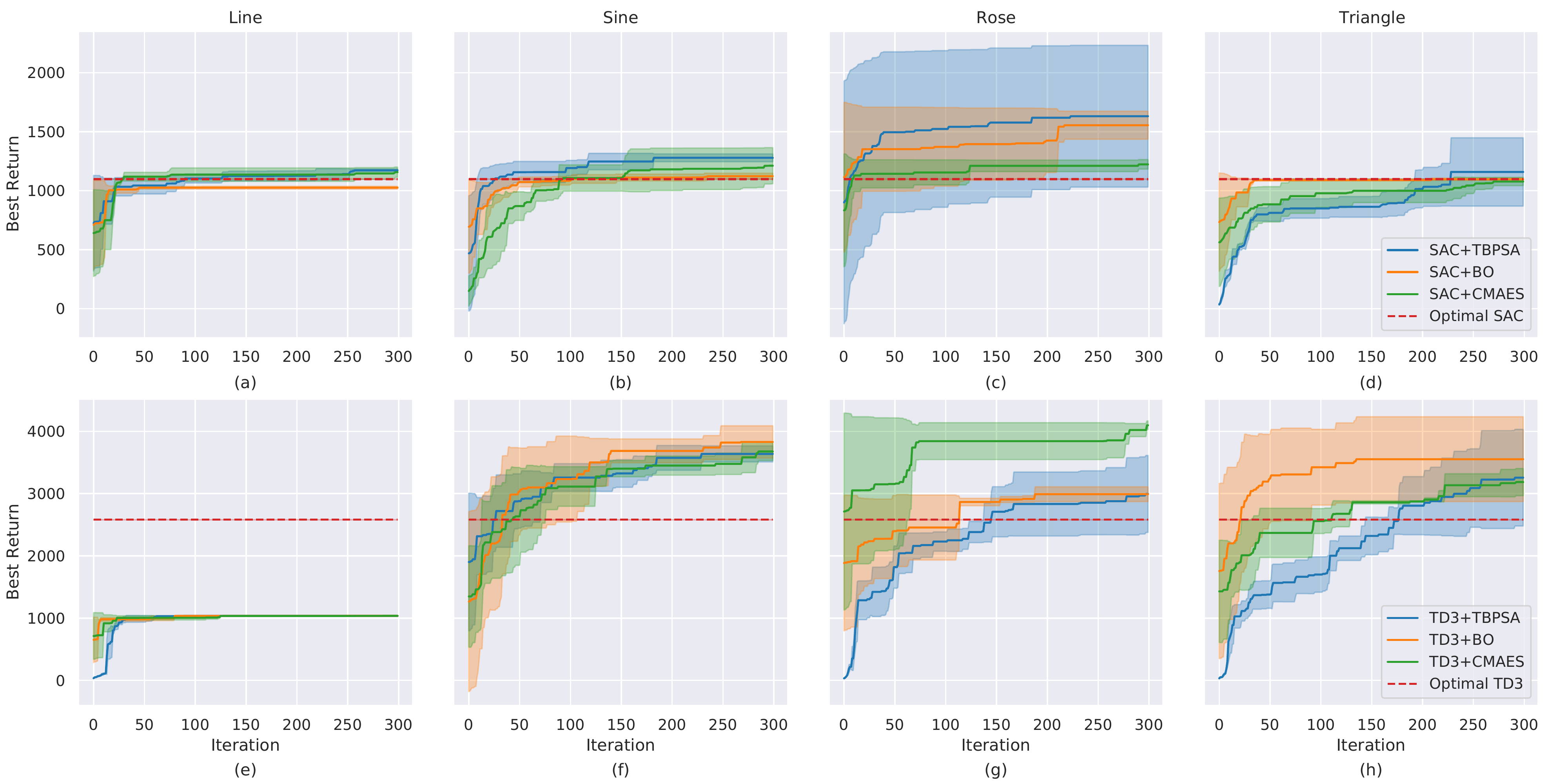}

     }}
        \vfill
        \caption{\textbf{Learning of bipedal locomotion using Automated Residual Reinforcement Learning (ARRL).} We show the training curves of our robot in bipedal mode trained with different algorithms based on Line, Sine, Rose, and Triangle gait patterns (motion primitives). For all these gaits, we show the ARRL combinations involving SAC (the first row) and involving TD3 (the second row).
        }
        \label{fig:arrl}
\end{figure*}

Fig. \ref{fig:arrl} shows the training curves of our ARRL algorithm with different combinations of RL agents (SAC or TD3) and black-box optimisers (TBPSA, CMAES, or BO). 
The first row shows the results of combinations that involve SAC and the highest scores of pure SAC; the second row shows the combinations that involve TD3 and the highest scores of pure TD3.


For all cases except methods involving TD3 with the Line gait (Fig. \ref{fig:arrl}e), all six ARRL combinations have a better performance than their corresponding underlying pure RL. For motion primitives of Sine, Rose, and Triangle, ARRL combinations involving TD3 generally have a better performance than combinations involving SAC and pure block-box optimisers. Among all the methods, the combination of TD3 and CMAES with the Rose gait is the best, reaching a return at around $4000$.



In terms of motion primitives, the design of the gait patterns is essential for this task. 
Methods involving the Line gait designs have similar performances which are much worse than that of those using other gait patterns. This reveals the dependency between the effectiveness of ARRL and the design of motion primitives. Among the Sine, Rose, and Triangle gaits, the highest average return occurs in the Rose gait.

The performance of our hybrid methods also heavily depends on the choice of RL agents. Fig. \ref{fig:rl} shows that the training of SAC is stuck in a local maximum in this task. This happens when SAC not only tries to optimise the discounted accumulated rewards but also the entropy of the policy, which adds too much noise to the optimisation of the Bellman equation (\ref{eq:bellman}) in this specific task setting. Therefore, the combination of SAC policy and parametric policy $\pi_{\theta^{\prime}}$ also does not work well in our task. On the other hand, the combination of TD3 policy and parametric policy $\pi_{\theta^{\prime}}$ shows the superiority of the ARRL frame. 

In terms of parameter optimisers, all of them can synergise with the learning process of RL agents within the ARRL framework. For three standard optimisers, it is shown that vanilla CMAES and TBPSA have a better performance than standard BO in this task. However, while they are combined with RL agents, TD3+BO outperforms the other two for Sine and Triangle gaits. This shows that the Gaussian process behind BO can manage the exploration-exploitation trade-off well during its cooperation with RL training.

Overall, standard TD3 can outperform SAC and all the parametric controllers with parameter optimisers, while our proposed ARRL with TD3 can even surpass the standard TD3. 
\subsection{Sim-to-Real Transfer}

\begin{figure}[tb]
     \centering
     \framebox{\parbox{0.47\textwidth}{
       \centering
     \begin{subfigure}[b]{0.11\textwidth}
         \centering
         \includegraphics[width=\textwidth]{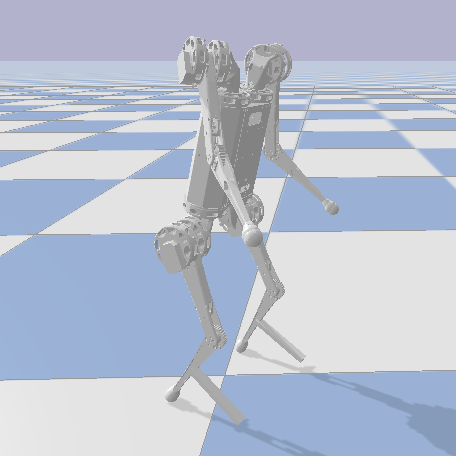}
     \end{subfigure}
     \hfill
     \begin{subfigure}[b]{0.11\textwidth}
         \centering
         \includegraphics[width=\textwidth]{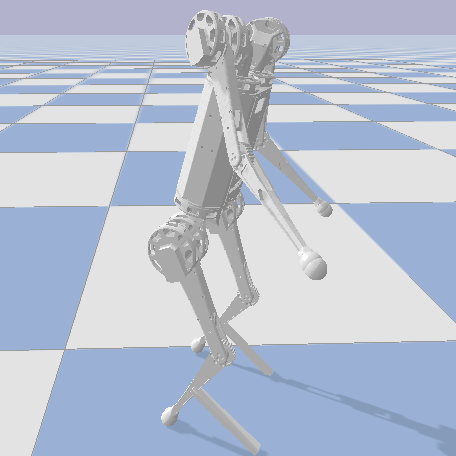}
     \end{subfigure}
     \hfill
     \begin{subfigure}[b]{0.11\textwidth}
         \centering
         \includegraphics[width=\textwidth]{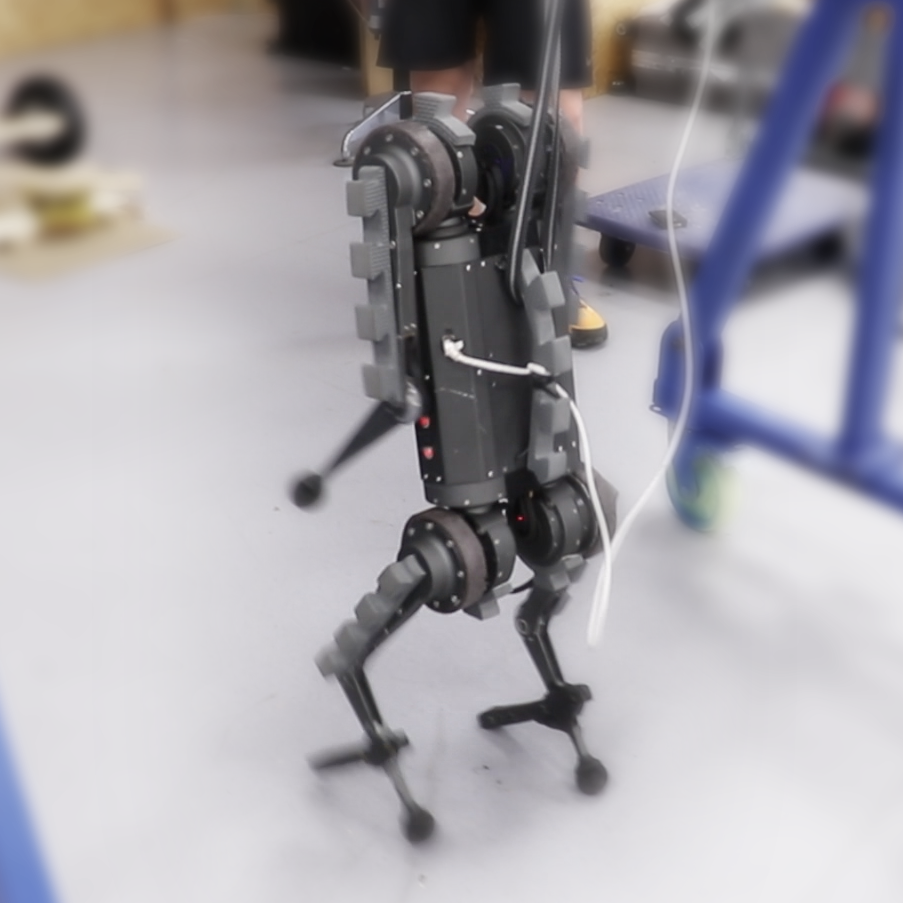}
     \end{subfigure}
     \hfill
     \begin{subfigure}[b]{0.11\textwidth}  
         \centering
         \includegraphics[width=\textwidth]{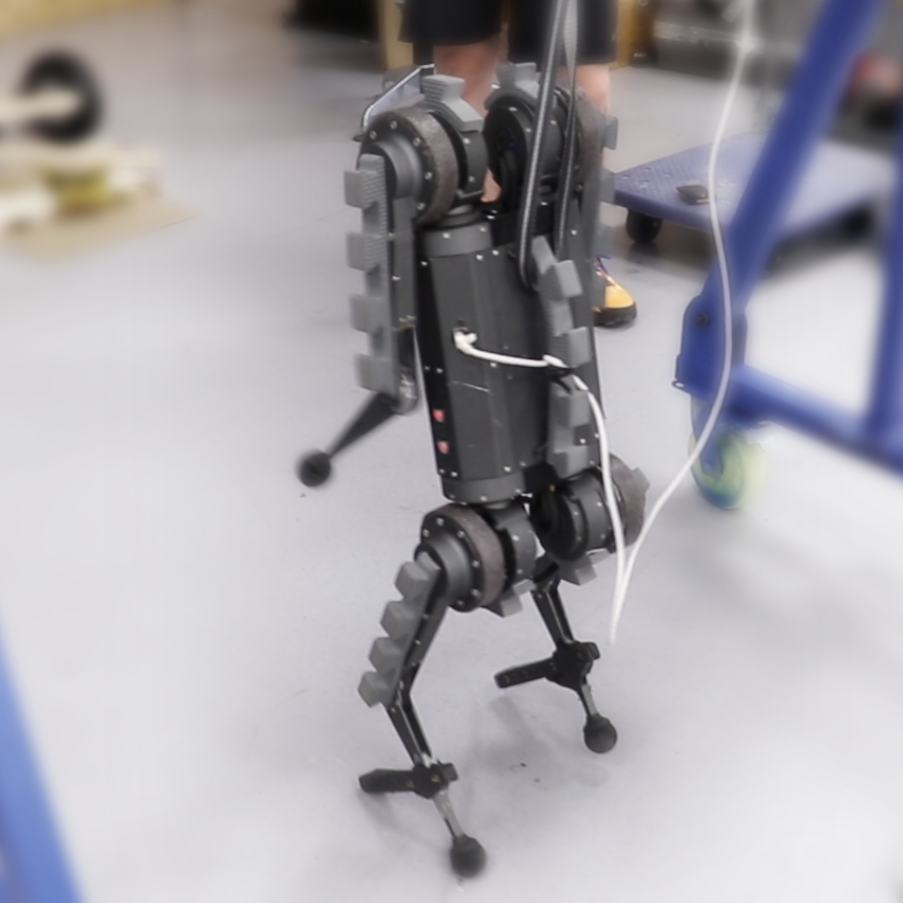}
     \end{subfigure}
     }}
        \caption{\textbf{Snapshots of the bipedal walking.} The robot performs a trained policy in the simulation (left) and real world (right).}
        \label{fig:walking}
\end{figure}


Firstly, we implement the policies trained by each method directly on the real robot. Snapshots of the walking of simulated and real robots are shown in Fig. \ref{fig:walking}. Returns of each algorithm are shown in Tab. \ref{tab:real_bad} with the same scoring metrics (accumulation of (\ref{eq:reward})) as in the simulation (also averaged over three trials). We also show the absolute walking distance of the robot in each case before the episode is finished---either the robot reaches the maximum episode length (1000 time steps, or 20 seconds) or falls down (the $e_{pitch}$ in (\ref{eq:error}) $ > 0.65$). We highlight the results with a walking distance greater than 1 meter with bold font.

The results show a significant reality gap, especially in the case of pure RL. For instance, while the pure TD3 agent in the simulator can obtain a return of $2580.14$, it can only reach a return of $31.23$ (walk for only 0.04 m) in the real world. 

We then implement the sim-to-real strategy described in Sec. \ref{sec:simtoreal} on the robot. The values of $\mathbf{e}_{\mathrm{threshold}}$ in (\ref{eq:prog0}, \ref{eq:prog1}) in our experiment are 0.15 and 0.35 for $e_{\mathrm{pitch}}$ and $\frac{d e_{\mathrm{pitch}}}{d t}$. Returns of each algorithm are shown in Tab. \ref{tab:real_good}. 

The results show that ARRL outperforms pure RL methods (Vanilla TD3 and Vanilla SAC) on real-world implementation, especially after applying the progressive sim-to-real strategy: four ARRL combinations reach a walking distance greater than 1 meter, while the majority has a walking distance better than pure RL methods. Although ARRL is not the best after the sim-to-real transfer---pure black-box optimisers (Vanilla TBPSA, BO, and CMAES) were capable of reaching the highest scores---it still shows a very good sim-to-real transferability.



As for the comparison among different motion primitives, the Triangle gait shows its advantage in the sim-to-real transfer. This can be explained by the design principle of the Triangle gait where each foot will first leave the floor horizontally to the ground. This not only creates a margin for positional error, but also allows its deformation without affecting the contact position between the feet and the floor.

\begin{table}[]
\caption{Accumulated rewards of the real robot and the corresponding bipedal walking distances (shown in the parentheses) with direct policy implementation.}
\begin{center}
\begin{tabular}{|c||c||c||c||c|}
\hline
& Sine & Rose & Triangle & Line\\
\hline
 SAC+TBPSA & 56.87 & 24.1 & 34.27 & 78.29\\
       & (\textbf{1.10 m}) & (0.16 m) & (0.11 m) & (0.41 m) \\
 SAC+BO & 30.1 & 32.48 & 22.03 & 48.24\\
       & (0.09 m) & (0.01 m) & (0.14 m) & (0.11 m) \\
 SAC+CMAES & 22.54 & 40.83 & 32.5 & 33.7\\
       & (0.09 m) & (0.31 m) & (0.15 m) & (0.23 m) \\
 TD3+TBPSA & 30.35 & 29.81 & 21.72 & 40.54\\
       & (0.59 m) & (0.15 m) & (0.25 m) & (0.25 m) \\
 TD3+BO & 22.49 & 25.28 & 20.45 & 28.96\\
       & (0.74 m) & (0.08 m) & (0.06 m) & (0.15 m) \\
 TD3+CMAES & 32.72 & 30.77 & 29.07 & 22.01\\
       & (0.35 m) & (0.45 m) & (0.18 m) & (0.14 m) \\
 Vanilla TBPSA & 351.51 & 86.85 & 57.03 & 786.87\\
       & (\textbf{1.79 m}) & (0.01 m) & (0.26 m) & (0.67 m) \\
 Vanilla BO & 473.95 & 511.02 & 350.8 & 760.31\\
       & (\textbf{1.90 m}) & (0.47 m) & (0.45 m) & (0.22 m) \\
 Vanilla CMAES & 527.75 & 398.56 & 80.95 & 780.43\\
       & (\textbf{1.19 m}) & (0.96 m) & (0.87 m) & (0.78 m) \\
\hline
 Vanilla SAC  & \multicolumn{4}{c|}{ 31.9 } \\
 & \multicolumn{4}{c|}{(0.01 m)} \\
 Vanilla TD3  & \multicolumn{4}{c|}{ 31.23 } \\
 & \multicolumn{4}{c|}{(0.04 m)} \\
\hline
\end{tabular}
\end{center}
\label{tab:real_bad}
\end{table}

\begin{table}[]
\caption{Accumulated rewards and the corresponding bipedal walking distances (shown in the parentheses) with the proposed sim-to-real strategy. 
}
\begin{center}
\begin{tabular}{|c||c||c||c||c|}
\hline
& Sine & Rose & Triangle & Line \\
\hline
 SAC+TBPSA & 86.89 & 39.05 & 253.00 & 77.62 \\
       & (0.95 m) & (0.78 m) & (0.34 m) & (0.67 m) \\
 SAC+BO & 105.10 & 58.79 & 237.91 & 292.01 \\
       & (0.30 m) & (0.42 m) & (0.07 m) & (0.38 m) \\
 SAC+CMAES & 68.90 & 105.93 & 237.42 & 85.46 \\
       & (0.26 m) & (0.69 m) & (\textbf{1.61 m}) & (0.87 m) \\
 TD3+TBPSA & 70.10 & 62.83 & 269.79 & 83.93 \\
       & (0.24 m) & (0.11 m) & (0.98 m) & (0.53 m) \\
 TD3+BO & 74.40 & 64.32 & 188.63 & 103.72 \\
       & (0.60 m) & (\textbf{1.16 m}) & (\textbf{2.01 m}) & (0.33 m) \\
 TD3+CMAES & 89.93 & 71.32 & 269.92 & 62.63 \\
       & (0.17 m) & (0.52 m) & (\textbf{2.47 m}) & (0.44 m) \\
 Vanilla TBPSA & 622.86 & 464.04 & 692.26 & 785.08 \\
       & (\textbf{1.15 m}) & (\textbf{3.72 m}) & (\textbf{1.55 m}) & (0.91 m) \\
 Vanilla BO & 614.79 & 599.63 & 735.90 & 760.69 \\
       & (0.95 m) & (0.56 m) & (\textbf{2.43 m}) & (0.03 m) \\
 Vanilla CMAES & 592.07 & 872.04 & 661.76 & 797.75 \\
       & (\textbf{1.13 m}) & (\textbf{4.75 m}) & (\textbf{4.28 m}) & (\textbf{2.20 m}) \\
\hline
 Vanilla SAC  & \multicolumn{4}{c|}{ 59.95 } \\
 & \multicolumn{4}{c|}{(0.18 m)} \\
 Vanilla TD3  & \multicolumn{4}{c|}{ 55.31 } \\
 & \multicolumn{4}{c|}{(0.03 m)} \\
\hline
\end{tabular}
\end{center}
\label{tab:real_good}
\end{table}

\section{Discussion And Future Work}
The proposed multi-modal locomotion framework consists of two modules: a hand-engineered multi-modal transition strategy (Fig. \ref{fig:standing}) and a learning-based control method ARRL that enables the robot to walk in the bipedal mode (Fig. \ref{fig:walking}). These two modules are intertwined for the same purpose: endow any off-the-shelf quadruped robot with the ability to walk bipedally. 
The hand-crafted transition strategy provides a solution to build a multi-modal system with a simple mechanical modification, while the customised controller ARRL can efficiently learn to walk in the bipedal mode without an assumption of the dynamics model.
In comparison to other robots with multiple locomotion modes \cite{huang2018design, Kobayashi2013LocomotionSS}, a robot with this multi-modal locomotion framework can be lighter and have a lower cost. This will make such robots more attractive, being a step towards commercialising multi-locomotion robots and bringing them into our daily life.

As our future work, we list some example applications of robots with our multi-modal locomotion framework as follows: a) the robot could stand up in bipedal mode to climb an obstacle that is even higher than its quadruped height, which can greatly increase the adaptability of the robot in different terrains; b) it could use its forelimbs as manipulators for pick-and-place tasks in its bipedal mode, c) or it could use its forelimbs to open doors or press buttons, which can enable the robot to work better in a human-centered environment. Along with all these benefits, the robot could still enjoy a faster and more stable gait in its quadruped mode.

\section{Conclusion}
In this paper, we propose a multi-modal locomotion framework that consists of two parts: a hand-engineered multi-modal transition strategy and a learning-based controller ARRL that enables the robot to walk in the bipedal mode.
We apply a mechanical structure on an off-the-shelf quadruped robot to provide a supporting polygon for its bipedal mode. 
We design a learning-based controller ARRL that uses parameter optimisers (CMAES, TBPSA, or BO) during the training of a residual RL agent (SAC or TD3). 
Our results show that ARRL outperforms all the standard RL methods and standard parameter optimisers in simulation and it also shows a good performance when implemented in the real robot.
Overall, our proposed framework shows the possibility to endow arbitrary quadruped robots with the ability to walk bipedally. This is a step towards commercialising multi-locomotion robots and bringing them into our daily life.

\bibliographystyle{IEEEtran} 
\bibliography{ref}

\end{document}